\documentclass{article}
\usepackage{spconf,amsmath,graphicx,hyperref}
\usepackage{fancyhdr} 
\usepackage{algorithm}
\usepackage{algorithmic}

\usepackage{amsmath}
\usepackage{float}
\usepackage{caption}
\usepackage{makecell}
\usepackage{booktabs} 
\usepackage{amssymb} 
\usepackage{times}
\usepackage{helvet}
\usepackage{courier}
\usepackage{xcolor}

\title{KD-CVG: A Knowledge-Driven Approach for Creative Video Generation}
\name{\begin{tabular}{@{}c@{}}
Linkai Liu$^{\dagger}$ \qquad Wei Feng$^{\star}$ \qquad Xi Zhao$^{\star}$ \qquad Shen Zhang$^{\star}$ \qquad Xingye Chen$^{\star}$ \\
Zheng Zhang$^{\star}$ \qquad Jingjing Lv$^{\star}$ \qquad Junjie Shen$^{\star}$ \qquad Ching Law$^{\star}$ \qquad Yuchen Zhou$^{\dagger}$ \\
Zipeng Guo$^{\dagger}$ \qquad Chao Gou$^{\dagger}$$^{\ddagger}$
\end{tabular}\thanks{$^{\ddagger}$Corresponding authors}}

\address{$^{\dagger}$ Sun Yat-sen University\\
$^{\star}$ JD.COM }
\begin{document}
%
\maketitle
\begin{abstract}
Creative Generation (CG) leverages generative models to automatically produce advertising content that highlights product features, and it has been a significant focus of recent research. However, while CG has advanced considerably, most efforts have concentrated on generating advertising text and images, leaving Creative Video Generation (CVG) relatively underexplored. This gap is largely due to two major challenges faced by Text-to-Video (T2V) models: (a) \textbf{ambiguous semantic alignment}, where models struggle to accurately correlate product selling points with creative video content, and (b) \textbf{inadequate motion adaptability}, resulting in unrealistic movements and distortions. To address these challenges, we develop a comprehensive Advertising Creative Knowledge Base (ACKB) as a foundational resource and propose a knowledge-driven approach (KD-CVG) to overcome the knowledge limitations of existing models. KD-CVG consists of two primary modules: Semantic-Aware Retrieval (SAR) and Multimodal Knowledge Reference (MKR). SAR utilizes the semantic awareness of graph attention networks and reinforcement learning feedback to enhance the model's comprehension of the connections between selling points and creative videos. Building on this, MKR incorporates semantic and motion priors into the T2V model to address existing knowledge gaps. Extensive experiments have demonstrated KD-CVG's superior performance in achieving semantic alignment and motion adaptability, validating its effectiveness over other state-of-the-art methods. The code and dataset will be open source at \url{https://kdcvg.github.io/KDCVG/}.
\end{abstract}
\section{Introduction}
\label{sec:intro}
\begin{figure*}[t]
  \centering
   \includegraphics[width=0.85\linewidth]{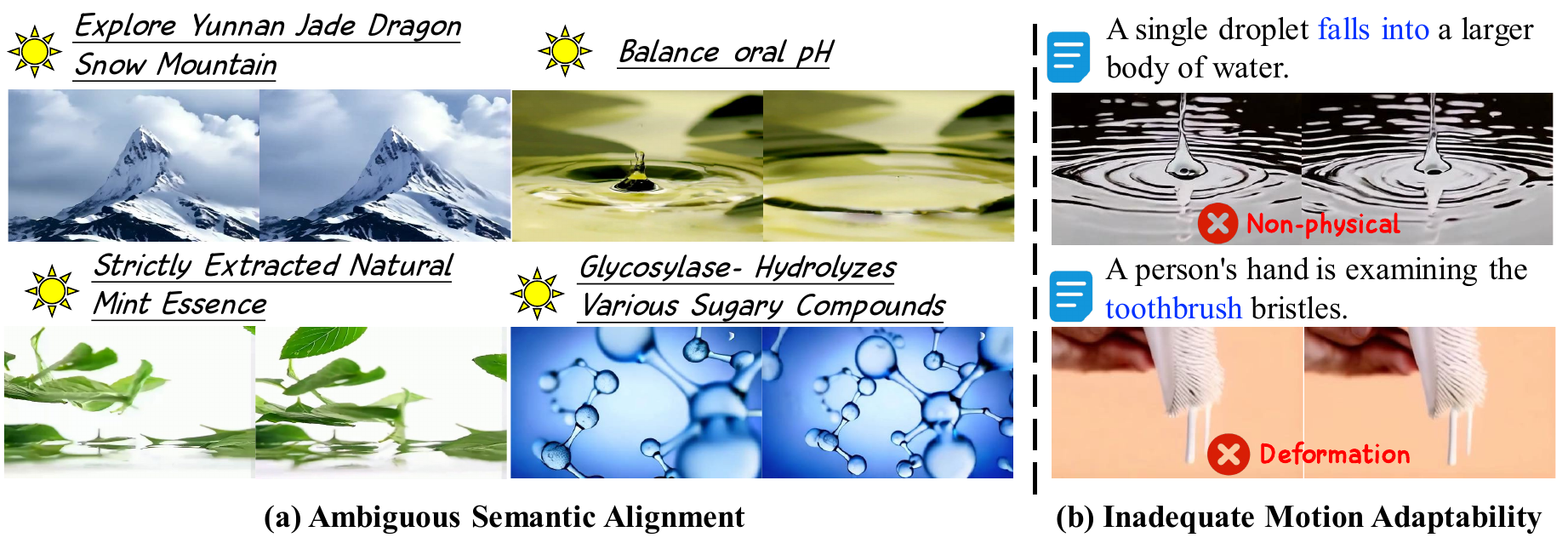}
   \caption{Illustration of challenges in T2V model generation of ACVs for E-commerce scenarios: (a) Low semantic correlation between selling point information and video content leads to alignment issues. (b) General T2V models, lacking e-commerce-specific motion priors, often display non-physical movements and motion deformations.}
   \label{fig:false}
\end{figure*}


Despite significant advancements in the Text-to-Video (T2V) field~\cite{du2024towards,yang2024new}, these technologies are not yet directly applicable to creative advertising video generation in e-commerce scenarios, primarily due to the following challenges:
(1) \textbf{Ambiguous Semantic Alignment.} Unlike general T2V tasks, CVG relies on selling point texts that are semantically ambiguous and often lack direct correlation with the video content. This makes it difficult for T2V models to capture the semantic relationship between the text and the video. As shown in Figure~\ref{fig:false} (a), Advertising Creative Videos (ACVs) use subtle motions of \textit{``water droplet"} and \textit{``mint leaves"} to indirectly convey the selling points of toothpaste, such as \textit{``Balance Oral pH"} and \textit{``Strictly Extracted Natural Mint Essence."}
(2) \textbf{Inadequate Motion Adaptability.} The effectiveness of ACVs lies in dynamically highlighting product features and unique selling points. Different products exhibit distinct movements based on their unique selling points, and these movements require a high degree of realism. Due to a lack of motion knowledge specific to e-commerce scenarios, general T2V models struggle to capture the complex and subtle motions in these contexts. This limitation frequently results in non-physical movements and distortions in the generated ACVs, as illustrated by the rising water droplets and distorted toothbrush in Figure~\ref{fig:false} (b).

To tackle the aforementioned challenges, we establish a comprehensive Advertising Creative Knowledge Base (ACKB) as a robust foundation for Creative Video Generation (CVG). Furthermore, we introduce a Knowledge-Driven Creative Video Generation approach (KD-CVG), aimed at leveraging the ACKB to fill the knowledge gaps in existing models. 
Our framework consists of two key modules:
\textbf{Semantic-Aware Retrieval (SAR)} addresses \textit{Ambiguous Semantic Alignment} in CVG through a Graph Attention Network (GAT) that identifies semantic relationships between texts. By retrieving relevant content from ACKB, SAR enhances semantic understanding of selling points and ensures text-video consistency.
\textbf{Multimodal Knowledge Reference (MKR)} leverages SAR's references to extract semantic and motion priors. Utilizing LLM's multimodal capabilities, MKR analyzes text-video connections and extracts dynamic patterns to generate physically plausible motions, effectively resolving \textit{Inadequate Motion Adaptation} issues.
The main contributions of this work are summarized as follows:

\begin{itemize}

\item We propose \textbf{KD-CVG}, a novel framework that automatically converts selling-point texts into ACVs, and release a comprehensive creative knowledge base to support future research.
\item We design the \textbf{SAR} module to retrieve semantically relevant references for the LLM, improving understanding of selling-point texts and mitigating ambiguous semantic alignment.
\item We introduce the \textbf{MKR} module, which leverages multimodal priors from references to produce motion-aligned ACVs, enhancing video quality and addressing motion adaptability challenges of T2V models in e-commerce.

\end{itemize}
\section{Method}

\begin{figure*}[t]
  \centering
   \includegraphics[width=0.9\linewidth]{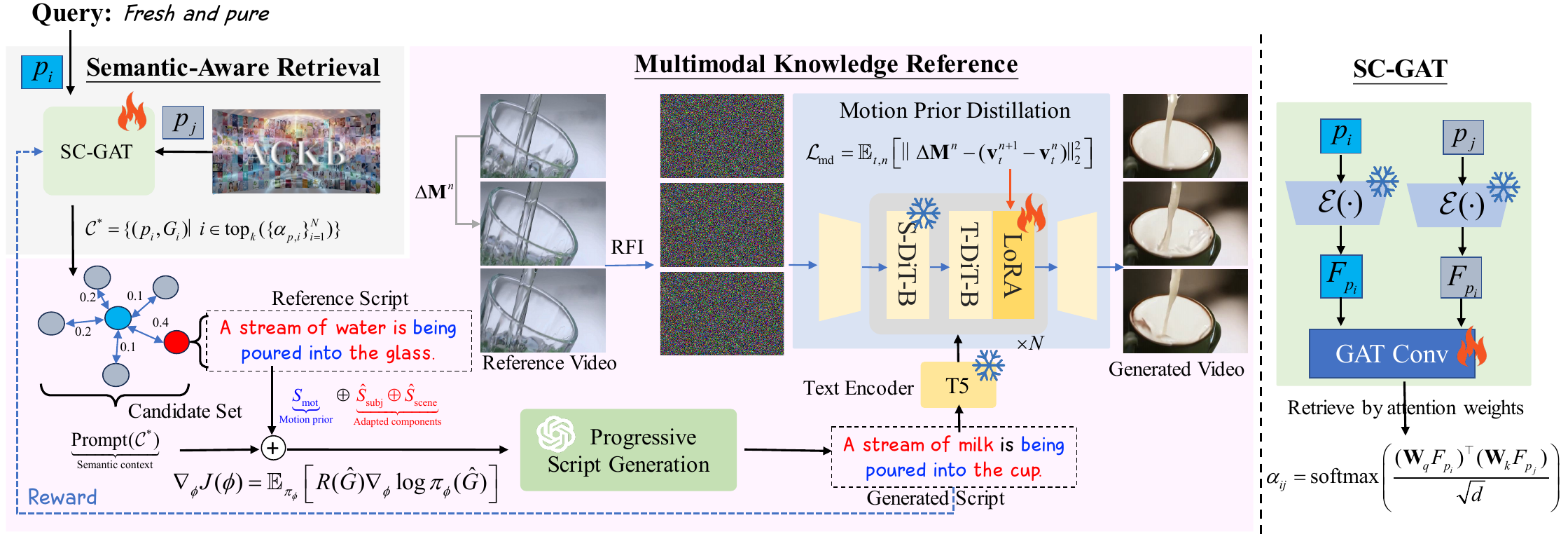}
   \caption{Overview of the two core modules in the KD-CVG framework.}
   \label{fig:kd-cavg}
\end{figure*}

\subsection{Advertising Creative Knowledge Base}
\begin{table}
\centering
\caption{Comparison of our dataset with other datasets.}
\resizebox{0.48\textwidth}{!}{\begin{tabular}{c|ccccc}
\hline
\textbf{Dataset} & \textbf{Domain} & \textbf{\makecell{Number of\\videos}} & \textbf{\makecell{Average\\duration}} & \textbf{\makecell{Average\\caption len}} & \textbf{Resolution} \\
\hline
MSVD~\cite{chen2011collecting} & Open & 1970 & $\sim$10s & 8.7 words & - \\
MSR-VTT~\cite{xu2016msr} & Open & 10K & $\sim$15s & 9.3 words & 240p \\
DiDeMo~\cite{anne2017localizing} & Flickr & 27K & $\sim$7s & 8.0 words & - \\
\hline
\textbf{ACKB (Ours)} & \makecell{E-commerce} & 10K & $\sim$3s & 8.8 words & 240p-2120p \\
\hline
\end{tabular}}
\label{tab:datainfo}
\end{table}
Our method, KD-CVG, relies on a high-quality Advertising Creative Knowledge Base (ACKB). We collected 58K ACVs from a major e-commerce platform, filtered low-quality videos following~\cite{qin2024xgen}, and used Qwen2-VL~\cite{Qwen2VL} to extract selling point texts. ProPainter~\cite{zhou2023propainter} removed watermarks and interfering text.
To address semantic misalignment between text and video, CogVLM2~\cite{hong2024cogvlm2} generated script descriptions for semantic references. Final quality control was conducted by experienced e-commerce reviewers. 
ACKB contains 10K text-video pairs covering 377 daily product types (e.g., toothbrushes, facial masks). Videos average 3 seconds with resolutions from 240p to 2120p. Challenges include limited data volume and short (avg. 8.8 words), vague selling texts, making ACKB a valuable but challenging resource for e-commerce video analysis (see Table~\ref{tab:datainfo}). 

\subsection{Semantic-Aware Retrieval}
\label{SAR}
To address semantic misalignment in CVG, we propose a Semantic-Aware Retrieval (SAR) module that establishes text-driven associations between product selling points and ACVs. As illustrated in Figure~\ref{fig:kd-cavg}, SAR progressively refine semantic alignment through graph-based reasoning and policy optimization.
The semantic space is formalized as a graph $\mathcal{G} = (\mathcal{V}, \mathcal{E})$ where nodes $\mathcal{V}$ represent selling point texts $\{p_i\}$ from the ACKB, and edges $\mathcal{E}$ encode their pairwise semantic relationships. Each node is initialized with CLIP-encoded text embeddings $F_{p_i} = \mathcal{E}(p_i)$, where $\mathcal{E}(\cdot)$ denotes the frozen CLIP text encoder. 
To model the semantic correlations among selling points, we introduce the Semantic Correlation Graph Attention Network (SC-GAT), which computes context-aware relevance scores through scaled dot-product attention:
\(    \alpha_{ij} = \text{softmax}\left(\frac{(\mathbf{W}_q F_{p_i})^\top (\mathbf{W}_k F_{p_j})}{\sqrt{d}}\right),\)
where $\mathbf{W}_q, \mathbf{W}_k \in \mathbb{R}^{d \times d}$ are learnable projection matrices, and $d$ denotes the embedding dimension. The attention weights $\alpha_{ij}$ quantify the semantic relevance between selling points $p_i$ and $p_j$, enabling adaptive aggregation of contextual information.


Given an input selling point $p$, the SC-GAT directly processes the CLIP-encoded text embeddings $F_p = \mathcal{E}(p)$ and $F_{p_i} = \mathcal{E}(p_i)$ to select top-$k$ candidates $\{(p_i, G_i)\}_{i=1}^k$, where $G_i$ denotes corresponding video script descriptions. The candidate selection process is optimized through policy gradient reinforcement learning~\cite{zhang2021sample}, where we maximize the expected CIDEr~\cite{vedantam2015cider} reward: \(
    \nabla_\phi J(\phi) = \mathbb{E}_{\pi_\phi} \left[ R(\hat{G}) \nabla_\phi \log \pi_\phi(\hat{G}) \right],  
\)
with $R(\hat{G}) = \text{CIDEr}(\hat{G}, G)$ representing the reward signal for the generated script $\hat{G}$ and ground truth $G$. The policy $\pi_\phi$ is parameterized by SC-GAT's attention weights $\phi$. 

\subsection{Multimodal Knowledge Reference}
\label{MKR}
Building upon the retrieval capability of SAR, our Multimodal Knowledge Reference (MKR) module establishes a hierarchical generation framework through two complementary stages: \textbf{progressive script generation} and \textbf{motion prior distillation}. This dual-stage architecture forms a closed-loop system that bridges semantic alignment from SAR with motion knowledge transfer, ensuring both semantic fidelity and physical plausibility in generated videos.

\subsubsection{Progressive Script Generation}
\label{subsec:psg}
Given an input selling point $p$, the SC-GAT computes the attention weights $\alpha_{p,i}$ between $p$ and all candidate selling points $\{p_i\}$ in the ACKB. The top-$k$ candidates are selected based on the highest attention weights \(\mathcal{C}^* = \{(p_i, G_i) \mid i \in \text{top}_k(\{\alpha_{p,i}\}_{i=1}^N)\}\), 
where $N$ is the total number of candidates in the ACKB. This selected set $\mathcal{C}^*$ is subsequently integrated into the prompt construction, serving as a semantic context to guide the LLM in generating the final script. 

\textbf{Step 1: Semantic Structure Parsing:} 
For the highest-ranked reference script $G^* \in \mathcal{C}^*$, we perform semantic decomposition through template-based parsing:
\(G^* \rightarrow \{S_{\text{subj}}, S_{\text{scene}}, S_{\text{mot}}\},\)
where $S_{\text{subj}}$ denotes symbolic subjects (e.g., ``mint leaves" representing freshness), $S_{\text{scene}}$ describes environmental contexts, and $S_{\text{mot}}$ captures motion patterns.

\textbf{Step 2: Motion-preserved Adaptation: }
We maintain the motion templates $S_{\text{mot}}^{(i)}$ while adapting subjects and scenes to target product features through constrained transformation:
\(\mathcal{T}(S_{\text{subj}}, S_{\text{scene}}) \rightarrow (\hat{S}_{\text{subj}}, \hat{S}_{\text{scene}}),\)
where $\mathcal{T}$ denotes the adaptation achieved through prompt engineering and contextual semantic reference of the candidate set.

\textbf{Step 3: Context-aware Synthesis: }  
The final script $\hat{G}$ integrates motion-preserved elements with semantic context from $\mathcal{C}^*$ through multimodal prompting:
\begin{equation}
\hat{G} = \text{LLM}\left(\underbrace{S_{\text{mot}}}_{\text{Motion prior}} \oplus \underbrace{\hat{S}_{\text{subj}} \oplus \hat{S}_{\text{scene}}}_{\text{Adapted components}} \oplus \underbrace{\text{Prompt}(\mathcal{C}^*)}_{\text{Semantic context}}\right).
\end{equation}


\subsubsection{Motion Prior Distillation}


Building upon temporal motion customization techniques~\cite{jeong2024vmc}, we establish a differential motion encoding scheme. Given a video's latent representation \( \mathbf{x}_t^{1:N} \in \mathbb{R}^{N \times d} \), where \( N \) denotes frame count and \( d \) the latent dimension, we compute frame-wise motion vectors as:
\(\mathbf{M}_t^n = \mathbf{x}_t^{n+1} - \mathbf{x}_t^n \quad \forall n \in \{1,...,N-1\}\).

Through linear interpolation in the latent space \( \mathbf{x}_t = t\mathbf{x}_1 + (1-t)\mathbf{x}_0 \), we derive continuous motion dynamics:
\(\mathbf{M}_t^n = t\mathbf{M}_1^n + (1-t)\mathbf{M}_0^n,\)
where \( \mathbf{M}_1^n \) and \( \mathbf{M}_0^n \) represent reference and initial motion vectors respectively. This formulation enables smooth motion interpolation across the temporal axis.
To align generated motion with reference patterns, we introduce a motion distillation loss that minimizes the discrepancy between target and predicted motion gradients:
\(    \mathcal{L}_{\text{md}} = \mathbb{E}_{t,n}\left[\|\Delta\mathbf{M}^n - (\mathbf{v}_t^{n+1} - \mathbf{v}_t^n)\|_2^2\right],\)
where \( \mathbf{v}_t^n \) denotes the learned velocity field in the rectified flow framework and \(\Delta\mathbf{M}^n = \mathbf{M}_1^n - \mathbf{M}_0^n\). This objective ensures temporal coherence by enforcing consistent motion transitions between consecutive frames.
For parameter-efficient adaptation, we integrate motion reference Low-Rank Adaptation~\cite{hu2022lora} (MR-LoRA) into the temporal attention blocks of the ST-DiT architecture. 

\section{Experiment}
\subsection{Implementation Details}
In our experiments, we use OpenSora v1.2~\cite{opensora} as the backbone and GPT-4 as the LLM. MR-LoRA is applied to the query projections in all self-attention layers of the T-DiT-B model with a rank of  $r = 128$. Training is conducted for 400 steps on a single NVIDIA H800 GPU using the Adam optimizer with a learning rate of  $1 \times 10^{-4}$ .
We evaluate our method on 27 selling-point texts from various categories (e.g., skincare and oral care). The reference videos contain diverse motion patterns such as water dripping, spin amplification, face washing, tooth brushing, and ocean waves, providing comprehensive scenarios to demonstrate performance.

\subsection{Quantitative and Qualitative Results}

\begin{table}
\renewcommand{\arraystretch}{0.85}
\centering
\caption{Comparison of the quality of scripts generated by LLM using different retrieval strategies.}
\resizebox{0.36\textwidth}{!}{
\begin{tabular}{c|cccc}
    \toprule
    \textbf{Methods} &\textbf{None} &\textbf{Random} & \textbf{COS} & \textbf{SC-GAT}  \\
    \midrule
    CIDEr & 17.81\% & 25.30\% &\underline{49.63\%} &\textbf{51.72\%} \\
    \bottomrule	
\end{tabular}}
\label{tab:sar}
\end{table}

\begin{figure}[t]
  \centering
   \includegraphics[width=0.98\linewidth]{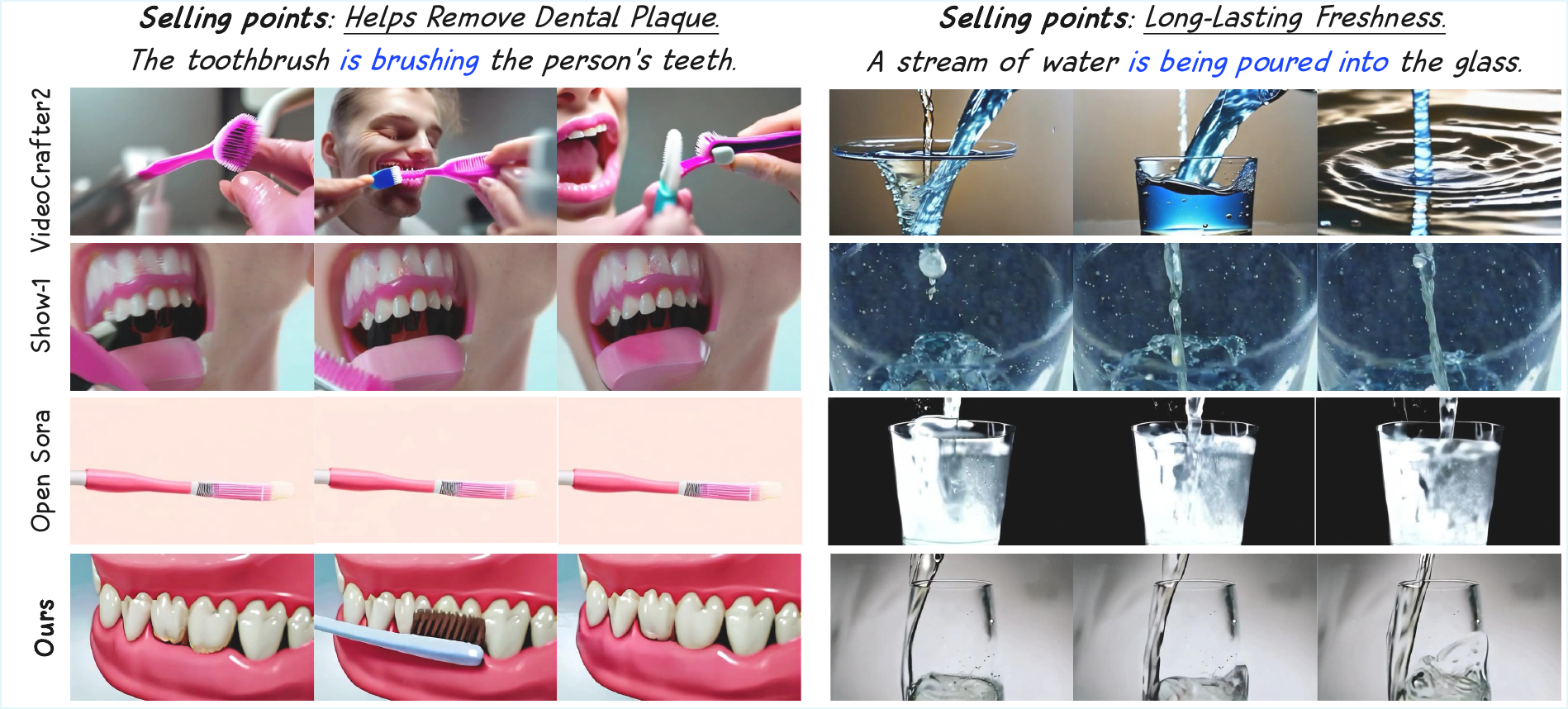}
   \caption{Qualitative comparison with other state-of-the-art video generation models.}
   \label{fig:comself}
\vspace{-12pt}
\end{figure}

\begin{table}
\renewcommand{\arraystretch}{0.85}
\centering
\vspace{-10pt}
\caption{Quantitative comparisons with SOTA models.}
\vspace{-10pt}
\resizebox{0.48\textwidth}{!}{
\begin{tabular}{c|cccc|c}
    \toprule
    \textbf{Methods} &\textbf{\makecell{Textual\\Alignment}} &\textbf{\makecell{Temporal\\Consistency}} & \textbf{\makecell{Dynamic\\Degree}} & \textbf{\makecell{Motion\\Smoothness}} & \textbf{\makecell{Min-Max\\Score}}  \\
    \midrule
    Show-1~\cite{zhang2023show} & \textbf{31.34\%} & 96.83\% & 44.44\% & 98.04\% & 55.66\%  \\
    VideoCrafter2~\cite{chen2024videocrafter2} & 30.43\% & 97.03\% & \textbf{88.89\%} & 91.91\% & 44.69\%  \\
    Open Sora~\cite{opensora} &28.84\% &97.94\% & 22.22\% &98.92\% & 47.06\% \\
    \midrule
    w/. MR-LoRA & 29.45\% & \textbf{98.15\%} &16.00\% &\textbf{99.24\%} & 56.10\% \\
    w/. RFI & 30.59\% & 97.55\% & 60.00\% & 97.52\% & \underline{65.36\%} \\
    \textbf{KD-CVG (Ours)} & \underline{30.68\%} & \underline{97.95\%} &\underline{72.00\%} &\underline{98.65\%} & \textbf{81.80\%} \\
    \bottomrule	
\end{tabular}}
\label{table2}
\end{table}

Figure~\ref{fig:comself} presents a qualitative comparison between our method and other state-of-the-art (SOTA) video generation models. It is evident that our method produces videos with superior motion realism and smoothness compared to other models. We quantitatively evaluate our method using automated metrics in Table~\ref{table2}. \textit{Textual Alignment} is the average CLIP similarity between all frame and text features; \textit{Temporal Consistency} measures CLIP similarity between consecutive frames; \textit{Dynamic Degree} and \textit{Motion Smoothness} are from VBench~\cite{huang2024vbench}. The \textit{Min–Max Score} averages normalized metrics for overall performance.
Compared to the baseline (Open Sora), MR-LoRA improves motion smoothness and temporal consistency but reduces dynamism. Adding Rectified Flow Inversion (RFI)~\cite{rout2024semantic} increases dynamism by 37.78\% at a slight cost to consistency and smoothness. Our full method (MR-LoRA + RFI) achieves the highest overall score.
User studies (Table~\ref{tab:time}) rate \textit{Text Alignment}, \textit{Video Quality}, \textit{Motion Regularity}, and \textit{Highlight Degree} on a 1--5 scale (27 participants). VideoCrafter2 scores higher in \textit{Video Quality} but performs poorly in \textit{Motion Regularity} (2.26 vs. 3.81). Our method improves substantially across all metrics.
KD-CVG inference time is 102~s, lower than Show-1 (3692~s) and VideoCrafter2 (161~s), and slightly higher than Open Sora (52~s), but with significantly better generation quality.


\begin{table}
\renewcommand{\arraystretch}{0.85}
\centering
\caption{Time cost and User study.}
\setlength{\tabcolsep}{0.05cm}
\resizebox{0.45\textwidth}{!}{
\begin{tabular}{c|cccc|c}
    \toprule
    \textbf{Methods} &\textbf{\makecell{Text\\Alignment}}$\uparrow$ &\textbf{\makecell{Video\\Quality}}$\uparrow$ & \textbf{\makecell{Motion\\Regularity}}$\uparrow$ & \textbf{\makecell{Highlight\\Degree}}$\uparrow$ & \textbf{\makecell{Time\\Cost}}$\downarrow$  \\
    \midrule
    Show-1~\cite{zhang2023show} & \underline{3.70} & 3.23 & \underline{3.44} & \underline{3.59} & 3692s  \\
    VideoCrafter2~\cite{chen2024videocrafter2} & 3.63 & \textbf{3.81} & 2.26 & 3.26 & 161s  \\
    Open Sora~\cite{opensora} &3.26 &3.07 & 2.63 &3.30 &\textbf{52s} \\
    \textbf{KD-CVG (Ours)} & \textbf{3.74} & \underline{3.70} &\textbf{3.81} &\textbf{3.89} & \underline{102s} \\
    \bottomrule	
\end{tabular}
}

\vspace{-10pt}
\label{tab:time}
\end{table}
\subsection{Ablation Studies}
To evaluate the SAR module's effect on semantic understanding of selling points, we compared CIDEr scores between model-generated and real scripts for 64 samples under different retrieval settings: `None' (no reference), `Random' (random references), and `COS' (cosine-similarity-based retrieval). Results in Table~\ref{tab:sar} confirm that SC-GAT significantly enhances semantic interpretation and coherence of generated scripts.
Table~\ref{tab:ablation} shows the ablation study assessing each component's contribution using automated metrics and user studies. Additionally, Table~\ref{tab:caption} examines how script quality (measured by CIDEr score) affects video generation, revealing a positive correlation between script quality and final video performance.

\begin{table}[h]
\vspace{-10pt} 
\renewcommand{\arraystretch}{0.75}
\centering
\hspace{0.005cm} 
\begin{minipage}{0.23\textwidth}
    \centering
    \caption{Ablation studies of different components.}
    \setlength{\tabcolsep}{0.02cm}
    \resizebox{0.99\textwidth}{!}{
    \begin{tabular}{c|c|c|cc}
        \toprule
        \textbf{Base} &\textbf{SAR} &\textbf{MKR} & \textbf{\makecell{Automatic\\Metrics}}$\uparrow$ & \textbf{\makecell{User\\Study}}$\uparrow$  \\
        \midrule
        \checkmark & &  &33.33\% &2.56  \\
        \checkmark &\checkmark & &\underline{53.54\%} &\underline{3.00}  \\
        \checkmark & &\checkmark &39.26\% &2.68 \\
        \checkmark &\checkmark &\checkmark &\textbf{83.12\%} &\textbf{3.80} \\
        \bottomrule	
    \end{tabular}
    }
    \label{tab:ablation}
\end{minipage}%
\hspace{0.005cm} 
\begin{minipage}{0.23\textwidth}
    \centering
    \caption{The impact of script generation quality on video generation performance.}
    \setlength{\tabcolsep}{0.02cm}
    \resizebox{0.99\textwidth}{!}{
    \begin{tabular}{c|c|cc}
        \toprule
        \textbf{\makecell{Script\\quality}} &\textbf{\makecell{Average\\CIDEr}}$\uparrow$  & \textbf{\makecell{Automatic\\Metrics}}$\uparrow$ &\textbf{\makecell{User\\Study}}$\uparrow$  \\
        \midrule
        Low &18.45\% &33.23\% &2.63  \\
        Medium &38.52\% &\underline{33.33\%} & \underline{2.75} \\
        High &87.95\% &\textbf{59.81\%} & \textbf{3.09} \\
        \bottomrule	
    \end{tabular}
    }
    \label{tab:caption}
\end{minipage}
\vspace{-10pt}
\end{table}

\vspace{-10pt} 
\section{Conclusion}
Existing models struggle to capture semantic nuances and motion dynamics in e-commerce videos. We propose KD-CVG, the first framework generating ACVs directly from selling points using a multimodal knowledge base, GAT-based semantic alignment, and motion priors. Experiments show it outperforms baselines in semantic alignment and motion adaptability, proving highly effective for practical applications.
\vfill\pagebreak

\bibliographystyle{IEEEbib}
\bibliography{strings,refs}

\end{document}